\documentclass[10pt,twocolumn,letterpaper]{article}

\usepackage[pagenumbers]{cvpr} 

\usepackage[accsupp]{axessibility} 
\usepackage{graphicx}
\usepackage{amsmath}
\usepackage{amssymb}
\usepackage{booktabs}
\usepackage{setspace}

\usepackage{pifont} 
\newcommand{\cmark}{\ding{51}}%
\newcommand{\xmark}{\ding{55}}%

\usepackage{graphicx}
\usepackage{booktabs}
\usepackage{multirow}
\usepackage{makecell}
\usepackage{caption}
\usepackage{threeparttable}
\usepackage{algorithm} 
\usepackage{algorithmic} 
\usepackage{listings}
\usepackage{bm}
\usepackage[normalem]{ulem}
\useunder{\uline}{\ul}{}
\usepackage[dvipsnames]{colortbl}
\usepackage[dvipsnames]{xcolor} 
\usepackage[most]{tcolorbox}
\definecolor{Gray}{gray}{0.85}
\usepackage{hhline}
\def\eg{\textit{e.g.}}
\def\ie{\textit{i.e.}}

\definecolor{KleinBlue}{rgb}{0.0, 0.129, 0.7}

\newcommand{\squishlist}{
	\begin{list}{$\bullet$}
		{ \setlength{\itemsep}{0pt}
			\setlength{\parsep}{1pt}
			\setlength{\topsep}{1pt}
			\setlength{\partopsep}{0pt}
			\setlength{\leftmargin}{1.5em}
			\setlength{\labelwidth}{1em}
			\setlength{\labelsep}{0.5em} } }
\newcommand{\squishend}{\end{list}}

\newcommand{\finding}[2]{
    \vspace{-0.1cm}
    \begin{tcolorbox}[
        colback=white!90!gray,
        colframe=teal!60!black,
        arc=5pt,
        boxsep=5pt,
        left=1pt,
        right=1pt,
        top=2pt,
        bottom=2pt,
        boxrule=0.8pt,
        drop shadow=gray!0!white,
        enhanced jigsaw
    ]
    \vspace{-0.1cm}
        \noindent\textbf{\textit{Finding #1:}} #2
    \vspace{-0.1cm}
    \end{tcolorbox}
    \vspace{-0.1cm}
}

\definecolor{cvprblue}{rgb}{0.21,0.49,0.74}
\usepackage[pagebackref,breaklinks,colorlinks,allcolors=cvprblue]{hyperref}
\def\unitemploc{\mbox{TimeLoc}}

\def\paperID{*****} 
\def\confName{CVPR}
\def\confYear{2025}

\title{\unitemploc: A Unified End-to-End Framework for Precise Timestamp Localization in Long Videos}
\author{
Chen-Lin Zhang$^{1\dagger}$\thanks{Equal contribution.} \hspace{0.1cm}
Lin Sui$^{1}$\footnotemark[1] \hspace{0.1cm}
Shuming Liu$^{2}$\footnotemark[1]  \hspace{0.1cm}
Fangzhou Mu$^{3}$\footnotemark[1] \hspace{0.1cm}
Zhangcheng Wang$^{4}$\ \hspace{0.1cm}
Bernard Ghanem$^{2}$\ \hspace{0.1cm}
\and
$^{1}$ Moonshot AI
\quad $^{2}$KAUST
\quad $^{3}$NVIDIA
\quad $^{4}$4Paradigm Inc
}
\begin{document}

\maketitle
\footnotetext[2]{Corresponding author: Chen-Lin Zhang $<$zclnjucs@gmail.com$>$}

\begin{abstract}
Temporal localization in untrimmed videos, which aims to identify specific timestamps, is crucial for video understanding but remains challenging. This task encompasses several subtasks, including temporal action localization, temporal video grounding, moment retrieval, and generic event boundary detection. Existing methods in each subfield are typically designed for specific tasks and lack generalizability across domains. 
In this paper, we propose \textbf{TimeLoc}, a unified end-to-end framework for timestamp localization that can handle multiple tasks. First, our approach employs a simple yet effective one-stage localization model that supports text queries as input and multiple actions as output. Second, we jointly train the video encoder and localization model in an end-to-end manner. To efficiently process long videos, we introduce temporal chunking, enabling the handling of videos with over 30k frames. Third, we find that fine-tuning pre-trained text encoders with a multi-stage training strategy further enhances text-conditioned localization.
TimeLoc achieves state-of-the-art results across multiple benchmarks: +1.3\% and +1.9\% mAP over previous best methods on THUMOS14 and EPIC-Kitchens-100, +1.1\% on Kinetics-GEBD, +2.94\% mAP on QVHighlights, and significant improvements in temporal video grounding (+11.5\% on TACoS and +6.7\% on Charades-STA under R1@0.5). Our code and checkpoints will be released at \url{https://github.com/sming256/TimeLoc}.
\end{abstract}

\section{Introduction}
\label{sec:intro}

In video understanding, many tasks involve localizing actions (\ie, temporal action localization~\cite{liu2020tsi,zhang2022actionformer,Liu_2024_CVPR}), text descriptions (\ie, temporal sentence grounding~\cite{sigurdsson2016charades,TACoS_ACL_2013}), and event boundaries (\ie, generic event boundary detection~\cite{lin2019bmn,gebd2021iccv,efficientgebd2024acmmm}) by identifying timestamps within an untrimmed video. These \emph{timestamp localization} tasks have attracted growing interest for their many applications in intelligent personal assistants~\cite{lei2021assistsr}, healthcare~\cite{liang2024audio}, and human-robot interaction~\cite{krajnik2015s,zhou2024embodied}, yet present a significant challenge as they require ``finding needles in a haystack" through joint reasoning over spatial, temporal, and textual information.

Although recent advances have led to methods that excel in individual tasks (\eg,~\cite{zhang2022actionformer,mu2024snag}), a unified modeling approach that generalizes across the spectrum of timestamp localization tasks is currently lacking, despite their shared problem structure. Moreover, existing methods often train only the localization head(s) on pre-extracted, \emph{frozen} video features, as naively forwarding a full-sized video through a typically large video encoder results in impractical GPU memory usage during gradient updates. However, this prevents the video encoder from adapting to the downstream task, potentially harming model accuracy.

\begin{figure}[t]
    \centering
    \includegraphics[width=0.85\linewidth]{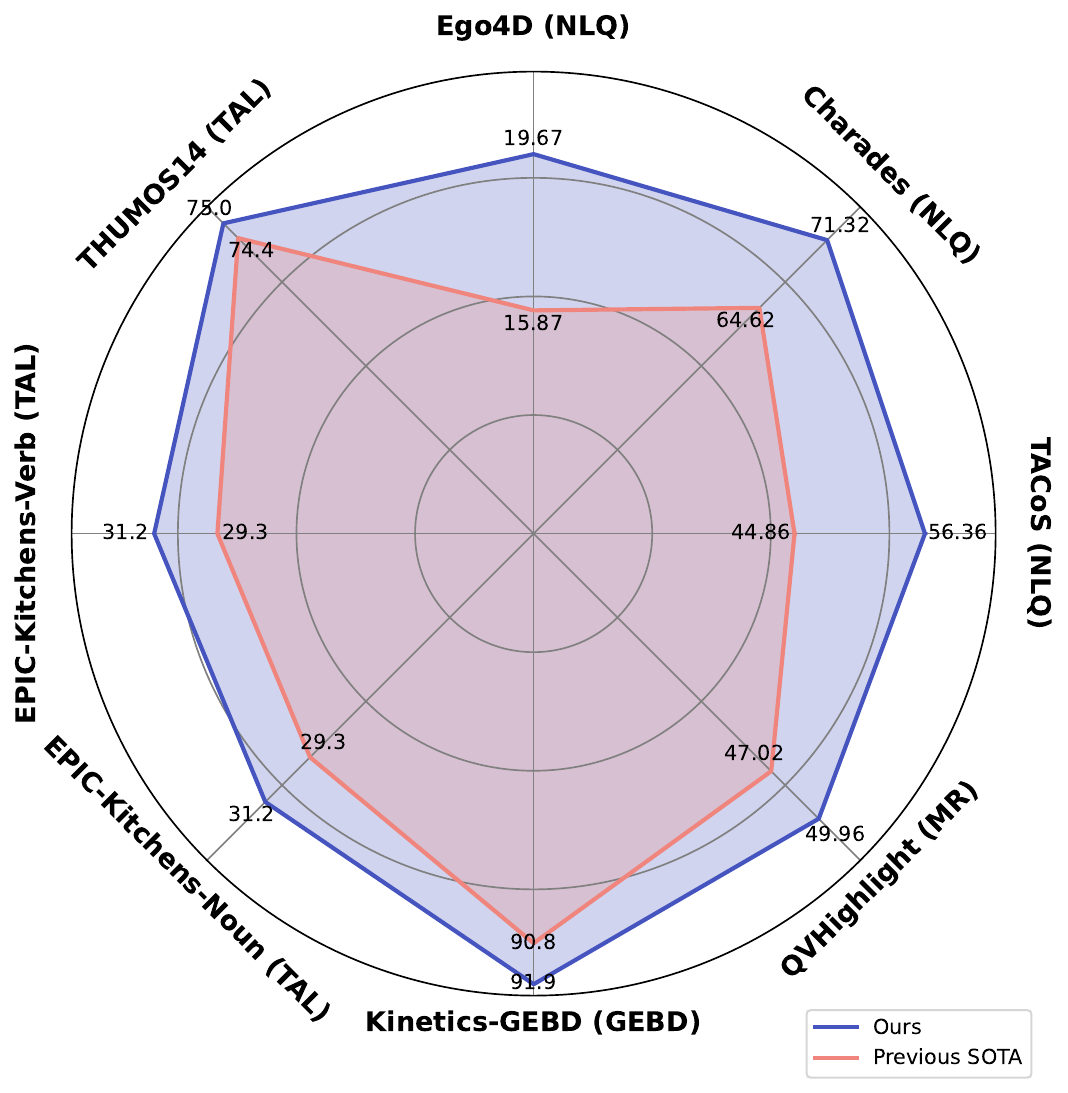}
    \vspace{-10pt}
    \caption{\textbf{TimeLoc achieves state-of-the-art performance across various temporal localization tasks}, including temporal action localization, temporal sentence grounding, moment retrieval, and generic event boundary detection.}
    \label{fig:intro}
    \vspace{-12pt}
\end{figure}

To overcome these challenges, we present \textbf{TimeLoc}, a unified framework for \underline{Time}stamp \underline{Loc}alization that supports efficient end-to-end training. TimeLoc follows a minimalist design that is both simple and effective. \emph{First}, compared to multi-stage approaches like UnLoc~\cite{yan2023unloc} and UniVTG~\cite{lin2023univtg}, TimeLoc features a lightweight one-stage model design that can tackle a broader range of timestamp localization tasks, including generic event boundary detection, while achieving stronger performance. 
\emph{Second}, given this universal model architecture, we develop a temporal chunking strategy to facilitate the \emph{joint training} of video encoder and localization head(s) on long-form videos, for the first time enabling the processing of over $30K$ frames on a single GPU.
\emph{Third}, on selected tasks with textual inputs, we demonstrate that fine-tuning the pre-trained text encoder following a multi-stage training strategy yields significant performance improvements.

Notwithstanding its simplicity, TimeLoc achieves state-of-the-art (SoTA) accuracy across numerous timestamp localization tasks, as summarized in Figure~\ref{fig:intro}. 
On \emph{temporal action localization}, it surpasses the previous SoTA method AdaTAD~\cite{Liu_2024_CVPR} by 1.3\% mAP on THUMOS14~\cite{jiang2014thumos} and 1.9\% mAP on EPIC-Kitchens-100~\cite{damen2018scaling}, using the same VideoMAE-L backbone. On \emph{generic event boundary detection}, TimeLoc outperforms EfficientGEBD~\cite{efficientgebd2024acmmm} by 1.1\% on Kinetics-GEBD~\cite{gebd2021iccv}. 
On \emph{temporal sentence grounding}, TimeLoc demonstrates a significant gain of 8.89\% R1@0.3 and 11.5\% R1@0.5 on TACoS~\cite{TACoS_ACL_2013}, and an edge of 6.7\% R1@0.5 and 6.7\% R1@0.7 on Charades-STA~\cite{sigurdsson2016charades}, when compared to SnAG~\cite{mu2024snag}. On \emph{text-based moment retrieval}, TimeLoc enables an average mAP improvement of 2.94\% on QVHighlights~\cite{lei2021detecting} over CG-DETR~\cite{moon2023correlation}.

\medskip 
\noindent \textbf{Our contributions} are summarized as follows:
\squishlist
\item We propose a unified, end-to-end framework for timestamp localization tasks including temporal action localization, temporal grounding, moment retrieval, and generic event boundary detection. Our framework can be easily extended to include other localization tasks.
\item We introduce temporal chunking to facilitate end-to-end training on long videos and find that fine-tuning the text encoder with a multi-stage training strategy further improves the accuracy of text-based timestamp prediction.
\item Our framework achieves state-of-the-art performance across a myriad of timestamp localization benchmarks, significantly outperforming previous methods.
\squishend

\section{Related work}
\label{sec:related work}

\subsection{Temporal Video Understanding}

\noindent\textbf{Temporal Action Localization} aims to identify action instances in untrimmed videos and classify their categories. Methods for this task can be divided into three types: one-stage, two-stage, and DETR-based approaches. One-stage methods, such as ActionFormer~\cite{zhang2022actionformer}, TriDet~\cite{shi2023tridet}, and DyFADet~\cite{yang2024dyfadet}, integrate action classification and boundary regression directly using multi-scale feature pyramids. Two-stage methods incorporate a proposal feature extraction step, exemplified by VSGN~\cite{zhao2021video}'s boundary sampling technique. Recently, end-to-end approaches have emerged~\cite{liu2023etad,Liu_2024_CVPR}. AFSD~\cite{lin2021learning} processes videos at a small spatial resolution for efficient end-to-end learning, while E2E-TAD~\cite{liu2022empirical} extensively studies end-to-end training strategies. 

\vspace{4pt}
\noindent\textbf{Generic Event Boundary Detection (GEBD)}, introduced by~\cite{gebd2021iccv}, focuses on detecting taxonomy-free event boundaries and segmenting videos into meaningful temporal chunks. Most existing works~\cite{ddmnet2022cvpr, efficientgebd2024acmmm} formulate GEBD as a binary classification problem. DDM-Net~\cite{ddmnet2022cvpr} constructs a feature bank to store temporal features and captures motion patterns using dense difference maps. EfficientGEBD~\cite{efficientgebd2024acmmm} establishes a strong baseline and utilizes video-domain backbones. A DETR-based detection model was proposed in~\cite{temporalperceiver2023pami} to address GEBD; however, it was trained offline. Existing GEBD methods are task-specific and lack generalizability across domains.

\vspace{4pt}
\noindent\textbf{Temporal Video Grounding} localizes moments in untrimmed videos based on text descriptions. Similar to temporal action localization, grounding methods are divided into two-stage and single-stage approaches, focusing on proposal generation and cross-modal fusion. Two-stage methods generate and score temporal segments, with some models conditioning on sentence queries to reduce dense sampling~\cite{xu2019qspn,chen2019sap,xiao2021bpnet,liu2021apgn,xiao2021lpnet}. In contrast, single-stage methods localize moments efficiently without explicit proposals, using global features or learnable queries~\cite{lei2021momentdetr,woo2022lvtr,liu2021cbln,zhao2021cpn,zhang2021seqpan}. Despite their efficiency, single-stage methods typically underperform compared to two-stage models on benchmarks. Cross-modal fusion integrates video and text information for improved grounding, evolving from simple late fusion to sophisticated early fusion strategies incorporating LSTMs, GCNs, and memory banks. Recent Transformer-based models~\cite{zhang2021matn,lei2021momentdetr,woo2022lvtr,barrios2023localizing, liu2024r} leverage early fusion by concatenating video features and text embeddings, enhancing cross-modal reasoning at the cost of increased model complexity. SnAG~\cite{mu2024snag} introduces a scalable single-stage framework that achieves state-of-the-art performance on major benchmarks. However, few works explore end-to-end temporal grounding, and scalability remains an open challenge.

\begin{figure*}[ht]
\centering
\vspace{-15pt}
\includegraphics[width=0.75\linewidth]{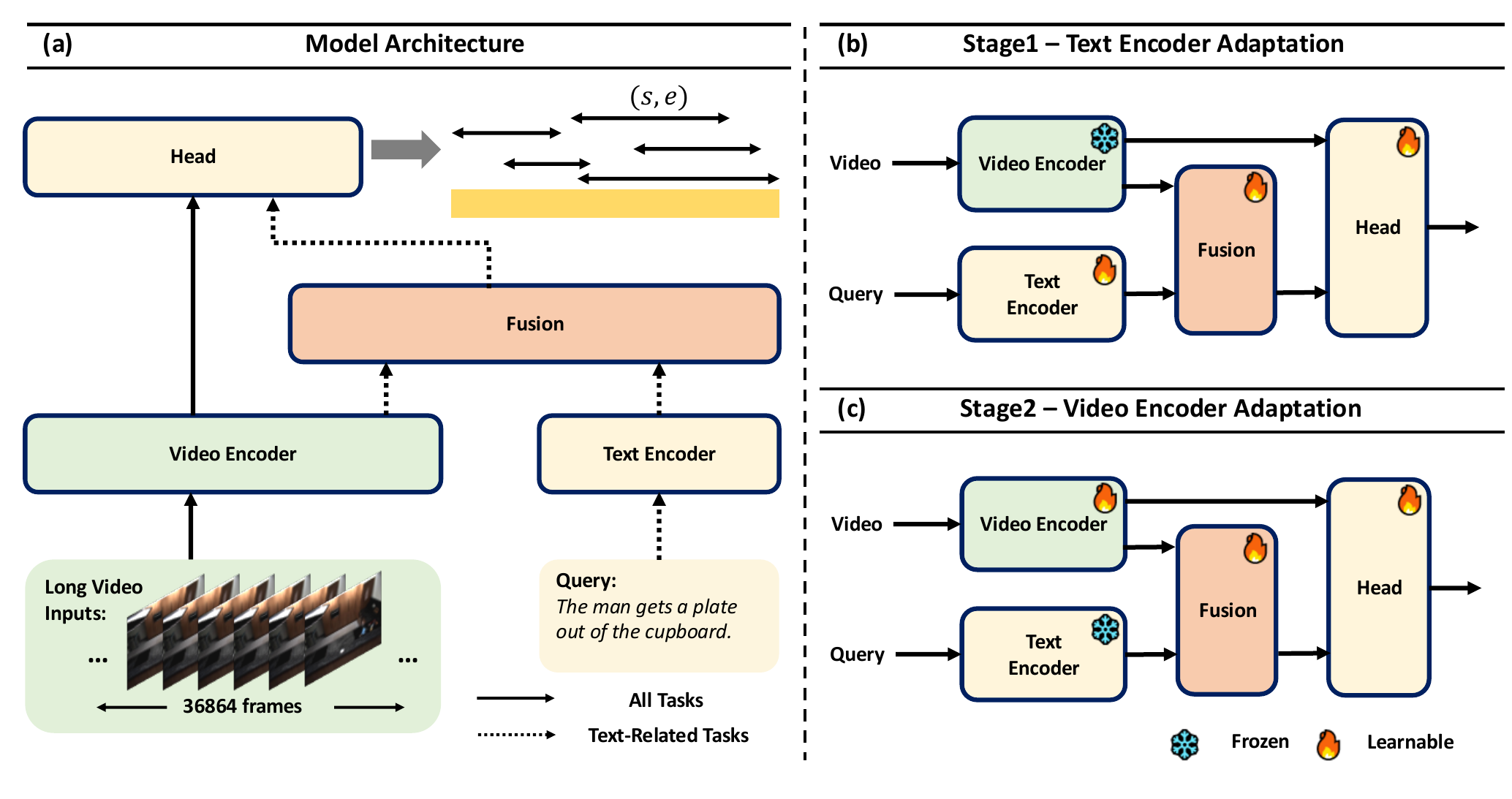}    
\vspace{-10pt}
\caption{\textbf{Pipeline and Multi-Stage Training Strategy of TimeLoc}. (a) The left diagram illustrates our pipeline, which supports temporal localization tasks with or without additional text conditions. The pipeline first extracts features from each modality, applies fusion strategies (if applicable), and then utilizes a lightweight localization head to generate predictions. On the right, the multi-stage training strategy is shown. The training process is divided into two stages: (b) in Stage 1, only the text encoder, fusion module, and localization head are trained. (c) In Stage 2, the text encoder is frozen while the video encoder, fusion module, and localization head are trained.}
\label{fig:method}
\vspace{-10pt}
\end{figure*}

\subsection{Efficient End-to-End Fine-tuning}

To reduce the cost of applying pre-trained video backbones to downstream tasks, researchers often use offline features as input~\cite{bmn2019iccv, zhang2022actionformer, mu2024snag}. AdaTAD~\cite{Liu_2024_CVPR} demonstrated the benefits of end-to-end (E2E) training for the temporal action localization task and introduced a temporal-informative adapter to reduce the number of trainable parameters. 
For data-intensive and computationally demanding video understanding tasks, E2E training consumes substantial device memory. To mitigate this, AdaTAD~\cite{Liu_2024_CVPR} implemented an alternative adapter placement policy to reduce memory consumption. While parameter-efficient fine-tuning (PEFT) strategies can alleviate memory constraints, they inherently limit the model's representational capacity.
In the development of large language models (LLMs), techniques such as Tensor Parallelism~\cite{megatron2019arxiv} and ZeRO~\cite{zero2020sc} have been employed to distribute resource consumption across multiple devices. 
\section{Methodology}

\subsection{Unifying Temporal Localization in Videos}

Given an untrimmed video $\mathbf{X} \in \mathbb{R}^{H \times W \times T \times 3}$, where $H$, $W$, and $T$ represent the height, width, and duration of the video, many video understanding tasks can be framed as predicting a set of segments ${(s_i, e_i, [y_i])}_{i=1}^{N}$, where $s_i$ and $e_i$ denote the start and end timestamps of an event and $y_i$ is an optional label. This formulation unifies different temporal localization tasks: 

\begin{itemize}
    \item \textbf{Temporal Action Localization} is a localization task where $y_i \in \{k_1,..., k_m\}$, a set of pre-defined actions with $m$ categories.
    \item \textbf{Temporal Video Grounding / Moment Retrieval} requires localizing video segments based on additional textual queries $\mathbf{Q}$. The query can be a natural language description of an action or a question requiring reasoning over the video content. In these settings, the $y_i$ is empty and the model only needs to generate timestamps.
    \item \textbf{Generic Event Boundary Detection} focuses on identifying temporal boundaries where events transition. We can use these event boundaries $\{b_0, b_1, ..., b_p\}$ to split the whole video into non-overlapping sequences where $b_0 = 0$ and $b_p$ is the length of the input video. Then the event boundary detection can be converted into timestamp predictions: $s_i=b_{i-1}$ and $e_i=b_i$.
\end{itemize}

Building upon the above formulation, the key to unifying these localization tasks lies in designing \textit{a generic localization backbone} that can effectively fuse video with other modalities, along with \textit{flexible localization heads} capable of generating dense timestamps and labels to accommodate varying localization objectives. To this end, we adapt SnAG~\cite{mu2024snag}, a one-stage, anchor-free approach as our localization framework. While SnAG was originally developed for video grounding only, we extend it to a broader range of temporal localization tasks with end-to-end training capability.

\subsection{One-Stage Localization Framework}

As illustrated in Figure~\ref{fig:method}, our one-stage localization framework consists of a video encoder, a text encoder, a fusion block, and a localization head. Please note that the text encoder and the fusion block are optional. Following~\cite{mu2024snag}, this framework maintains a simple yet effective design, enabling the prediction of multi-scale actions and accommodating videos of varying lengths, particularly long-form videos. For more details of the fusion block architecture, we refer the reader to~\cite{mu2024snag}. 

Next, we present the key modifications introduced in our framework, focusing on three main aspects: video / text feature extraction, the localization head, and the end-to-end training strategy. These enhancements allow the model to generalize across different temporal localization tasks while maintaining efficiency and scalability.

\vspace{4pt}
\noindent\textbf{Video Feature Extraction.} To encode video frames into compact feature representations, we employ a powerful off-the-shelf video encoder $\mathcal{F}$, \textit{i.e.,} VideoMAE~\cite{tong2022videomae}, though our approach is not limited to this specific model. VideoMAE is pretrained on large-scale video datasets using self-supervised learning followed by supervised fine-tuning, making it well-suited for robust feature extraction. 

Specifically, we discard the classification head of VideoMAE, and apply the spatial average pooling after the backbone to obtain the frame-level video representation $f_V \in \mathbb{R}^{C \times T}$, where $C$ is the feature dimension and $T$ is the temporal length. Then, $f_V$ is passed to the subsequent fusion block for temporal modeling with additional text features.

Since VideoMAE operates only on 16-frame inputs, we follow prior work~\cite{Liu_2024_CVPR} by dividing the video into non-overlapping 16-frame clips. Each clip is independently processed by the encoder, and the extracted features are subsequently aggregated to form the full video representation.

\vspace{4pt}
\noindent\textbf{Text Feature Extraction.} Similarly, given a text query, we use a text encoder $\mathcal{G}$ to extract the text feature $f_Q \in \mathbb{R}^{C \times L}$, where $L$ is the number of text tokens. In this paper, we explore CLIP and GloVe as text encoders. Then, the video feature $f_V$ and text feature $f_Q$ are fused together by the fusion block to produce the multi-scale feature $f_H$.

\vspace{4pt}
\noindent\textbf{Localization Head.} The localization head $\mathcal{H}$ takes as input the video-text fused feature $f_H$ and estimates the start and end offsets $(\Delta s_i, \Delta e_i)$, which represent the relative distances to anchor points or reference timestamps, following an anchor-free approach. It also predicts the confidence score $c_i$. This formulation enables the model to generalize across different temporal localization tasks while maintaining flexibility in handling task-specific objectives.

Unlike the standard video grounding head in SnAG, our approach also supports a multi-class classification objective, which is essential for temporal action localization tasks. 
Furthermore, it also supports saliency score prediction for highlight detection, extending the framework to a broader range of temporal localization tasks.

\vspace{4pt}
\noindent\textbf{End-to-End Training.}  
Beyond the aforementioned model design, we also jointly fine-tune the video encoder $\mathcal{F}$ along with the fusion block and localization head $\mathcal{H}$. End-to-end training helps bridge the gap between pretraining and fine-tuning by adapting the model to task-specific data distributions and mitigating domain discrepancies. Moreover, this approach enables the model to learn more robust video representations, reducing the risk of overfitting in the localization head. In our experiments, we find that end-to-end learning is crucial for precise temporal localization, and surprisingly, that a lightweight end-to-end backbone can outperform offline features extracted by gigantic video encoders.

\subsection{Handling Long Videos}
Our proposed framework effectively performs timestamp localization for most tasks. However, for extremely long videos (\textit{e.g.,} exceeding thirty minutes), we observe that GPU memory constraints arise, as all video frames must be loaded into memory for end-to-end learning. Detailed ablation studies are provided in Section~4. Previous temporal action localization methods, such as AdaTAD~\cite{Liu_2024_CVPR}, address this issue by freezing the entire backbone and employing ladder networks to reduce memory usage. However, fine-tuning only portions of the video backbone often leads to suboptimal performance and remains insufficient for handling a large number of input frames.

In TimeLoc, inspired by gradient checkpointing~\cite{chen2016training}, we introduce a temporal gradient checkpointing technique into our framework. First, we divide each input video $\mathbf{X} \in \mathbb{R}^{H \times W \times T \times 3}$ into $t$ small temporal chunks along the temporal dimension: Given a partition function $\pi: \{1,2,\ldots,K\} \rightarrow \{1,2,\ldots,t\}$ that maps each index of the third dimension to its corresponding partition, then $\mathcal{K}_p = \{k \in {1,2,\ldots,T\} : \pi(k) = p}$ representing the indices of the third dimension that belong to partition $p$, the divide process can be:
\begin{align}
   \mathbf{X}_t &= X_{:, :, \mathcal{K}_p,:}, \nonumber \\
   \mathbf{X} &= \bigcup_{i=1}^{t}\mathbf{X_{:, :, \mathcal{K}_p,:}}.
\end{align}

For simplicity, we just sequentially divide $\mathbf{X}$ into small chunks. Then, we perform individual forward passes for each chunk $\mathbf{X}_i$, where $1 \leq i \leq t$, and aggregate the output features into a complete sequence $f$ along the temporal dimension. Importantly, this temporal gradient checkpointing strategy preserves all gradient information, enabling lossless long-video training with reduced peak memory usage at the cost of additional computation time. This approach allows TimeLoc to process nearly arbitrary frame counts, for instance, videos exceeding 36,000 frames.

\subsection{Improving Text-Conditioned Timestamp Localization by Fine-Tuning the Text Encoder}

Until now, we have maintained the pre-trained text encoder $\mathcal{G}$ unaltered in our framework, with text features $f_Q$ remaining frozen throughout training. Notably, previous research has consistently employed fixed text encoders while tuning video encoders, a practice that warrants further investigation. In our early attempts, fine-tuning both the video and text backbones simultaneously resulted in performance degradation compared to fine-tuning the video backbone alone. This decline likely stems from the modality collision between textual and visual representations.

To address this challenge, we propose a multi-stage training strategy that balances efficiency and effectiveness, as shown in Figure~\ref{fig:method} (b) and (c):
\squishlist
\item \textbf{Stage 1}: Freeze the video backbone $\mathcal{F}$ and fine-tune the text encoder $\mathcal{G}$, the fusion block, and the localization head $\mathcal{H}$. 
This initial stage introduces negligible computational overhead to the offline training pipeline.
\item \textbf{Stage 2}: Freeze the text encoder $\mathcal{G}$, only fine-tune the video backbone $\mathcal{F}$ and remaining components.
\squishend

Our experiments show that this progressive fine-tuning approach effectively mitigates the modality collision problem, leading to improved performance.

\section{Experiments and Analysis}
\subsection{Experiment Settings}

\noindent\textbf{Datasets.}
We evaluate our proposed method on different localization tasks with the following datasets:
\squishlist
\item \textbf{THUMOS14}~\cite{jiang2014thumos} consists of 413 untrimmed videos spanning 20 action categories, with actions ranging from 0.2 to 118 seconds.
This dataset is widely used for the temporal action localization task.
\item \textbf{EPIC-Kitchens 100}~\cite{damen2018scaling} contains 700 egocentric videos with various actions. Actions in this dataset are defined by verb-noun combinations.
\item \textbf{TACoS}~\cite{TACoS_ACL_2013} is a temporal grounding benchmark consisting of 10.1 hours of cooking videos, with an average of 143.5 queries per video. The dataset exhibits significant variation in video lengths (ranging from 52 seconds to approximately 40 minutes).
\item \textbf{Charades-STA}~\cite{sigurdsson2016charades} consists of videos averaging 30 seconds in length, with each video associated with an average of 2.4 text queries.
\item  \textbf{QVHighlights}~\cite{lei2021momentdetr} includes 10,148 videos and 10,310 text queries. Each video has a maximum duration of 150 seconds. This dataset supports both moment retrieval and highlight detection tasks.
\item \textbf{Ego4D}~\cite{grauman2022ego4d} is a large-scale collection of egocentric videos capturing daily human activities, encompassing over 3000 hours of content. We conducted experiments with the Ego4D-v1 dataset.
\item \textbf{Kinetics-GEBD}~\cite{gebd2021iccv} is a generic event boundary dataset containing 54,691 videos randomly sampled from Kinetics-400~\cite{kay2017kinetics}, annotated with 1,290,000 generic event temporal boundaries.
\squishend

\noindent\textbf{Metrics.}
For temporal action localization tasks, we report the mean Average Precision (mAP) at different temporal Intersection over Union (tIoU) thresholds and the average mAP. Following ActionFormer~\cite{zhang2022actionformer}, we evaluate the predictions using tIoU thresholds of
$\{0.3, 0.4, 0.5, 0.6, 0.7\}$ on THUMOS14. For EPIC-Kitchens 100, the thresholds are set to $\{0.1, 0.2, 0.3, 0.4, 0.5\}$.

For temporal grounding tasks, we report Recall@$k$, where $k \in \{1, 5\}$, at various tIoU thresholds. On the TACoS dataset, the tIoU thresholds are set to $\{0.3, 0.5\}$, while for Charades-STA, we use thresholds of $\{0.5, 0.7\}$. 

For moment retrieval tasks, we report the mAP under tIoU of 0.5 and 0.75, as well as the average mAP with tIoU from 0.5 to 0.95 with 10 steps on QVHighlights.

For generic event boundary detection tasks, we follow prior work~\cite{gebd2021iccv, efficientgebd2024acmmm, ddmnet2022cvpr} and use the Relative Distance (Rel.Dis.) as the evaluation metric. Rel.Dis. quantifies the error between detected and ground truth timestamps, normalized by the length of the action instance. We report F1 scores with Rel.Dis. thresholds ranging from 0.05 to 0.5, as well as the average score for comparative analysis.

\subsection{Implementation Details}

All experiments are implemented using PyTorch 2.1. By default, we enable memory-efficient techniques such as automatic mixed-precision training and layer-wise gradient activation checkpointing (not to be confused with the proposed temporal gradient checkpointing). We use the AdamW optimizer~\cite{loshchilov2017fixing} with a weight decay of 0.05 for all experiments. Unless specifically mentioned, we use spatial resolution $160^2$ for VideoMAE, and CLIP~\cite{radford2021clip} as text encoder. For GEBD tasks, we convert the predicted action boundaries (start and end timestamps) into event boundaries.

For task-specific post-processing, we first apply non-maximum suppression (NMS) to all outputs. We select the top-$k$ predictions for most subtasks, including temporal action localization, temporal grounding, and moment retrieval. For GEBD tasks, we directly convert top prediction timestamps into event boundaries, discarding boundaries that are too close. Additional implementation details are provided in the supplementary materials.

\begin{table*}[t]
    \centering
    \scriptsize
    \caption{\textbf{Temporal Action Localization Results on THUMOS-14 and EPIC-Kitchens 100.} We report mAP\% at different tIoUs. E2E means end-to-end training, and Flow refers to offline extracted flow features. Best results are in \textbf{bold}, and the second-best results are \underline{underlined}. $^\dag$ means results were obtained under 224$^2$ spatial resolution.}
    \scriptsize
    \setlength{\tabcolsep}{2pt}
    \begin{tabular}{l|ccc|cccccc|cccccc|cccccc}
    \toprule
    \multirow{2}{*}{\textbf{Method}} & \multicolumn{1}{c}{\multirow{2}{*}{\textbf{Backbone}}} & \multicolumn{1}{c}{\multirow{2}{*}{\textbf{E2E}}} & \multirow{2}{*}{\textbf{Flow}} & \multicolumn{6}{c}{\textbf{THUMOS-14}}  & \multicolumn{6}{c}{\textbf{EPIC-Verb}} & \multicolumn{6}{c}{\textbf{EPIC-Noun}} \\ \cline{5-22} 
                       &               &            &                       & \textbf{0.3} & \textbf{0.4} & \textbf{0.5} & \textbf{0.6} & \textbf{0.7} & \textbf{Avg.} & \textbf{0.1} & \textbf{0.2} & \textbf{0.3} & \textbf{0.4} & \textbf{0.5} & \textbf{Avg.} & \textbf{0.1} & \textbf{0.2} & \textbf{0.3} & \textbf{0.4} & \textbf{0.4} & \textbf{Avg.} \\
    \hline
    ActionFormer~\cite{zhang2022actionformer} & SlowFast-R50 & \xmark & \xmark & 78.7 & 73.3 & 65.2 & 54.6 & 39.7 & 62.3 & 26.6 & 25.4 & 24.2 & 22.3 & 19.1 & 23.5 & 25.2 & 24.1 & 22.7 & 20.5 & 17.0 & 21.9 \\
    ASL~\cite{shao2023action}  & I3D &\xmark &\cmark& 83.1 & 79.0 & 71.7 & 59.7 & 45.8 & 67.9 & 27.9 & - & 25.5 & - & 19.8 & 24.6 & 26.0 & - & 23.4 & - & 17.7 & 22.6 \\
    TriDet~\cite{shi2023tridet} & I3D &\xmark &\cmark& 83.6 & 80.1 & 72.9 & 62.4 & 47.4 & 69.3 & 28.6 & 27.4 & 26.1 & 24.2 & 20.8 & 25.4 & 27.4 & 26.3 & 24.6 & 22.2 & 18.3 & 23.8 \\
    ActionFormer~\cite{zhang2022actionformer} &VideoMAE-L&\xmark&\cmark& - & - & - & - & - & - & 32.7 & 31.6 & 29.1 & 26.7 & \underline{23.6} & 28.7 & 31.3 & 29.7 & 27.2 & 25.3 & 21.3 & 26.9\\
    DyFADet~\cite{yang2024dyfadet}  & \scriptsize{VideoMAE-g} & \xmark & \cmark &85.4 & - & 74.0 & - & 50.2 & 71.1 & - & - & - & - & - & - & - & - & - & - & - & - \\ 
            
    \hline
    \hline

    AdaTAD~\cite{Liu_2024_CVPR} & SlowFast-R50 &\cmark &\xmark & - & - & - & - & - & - & 26.5 & 25.7 & 23.9 & 21.7 & 17.6 & 23.1 & 24.5 & 23.6 & 22.3 & 20.0 & 16.5 & 21.4 \\
    AdaTAD~\cite{Liu_2024_CVPR}$^\dag$ & VideoMAE-B & \cmark &\xmark & - & - & - & - & - & 71.9 & - & - & - & - & - & - & - & - & - & - & - & - \\
    AdaTAD~\cite{Liu_2024_CVPR}  & VideoMAE-L &\cmark &\xmark & 87.7 & 84.1 & 76.7 & 66.4 & 52.4 & 73.5 & \underline{33.1} & \underline{32.2} & \underline{30.4} & \underline{27.5} & 23.1 & \underline{29.3} & \underline{32.4} & \underline{31.6} & \underline{30.1} & \underline{27.4} & \underline{24.6} & \underline{29.3} \\
    AdaTAD~\cite{Liu_2024_CVPR}$^\dag$ & VideoMAE-L & \cmark &\xmark & - & - & - & - & - & 73.7 & - & - & - & - & - & - & - & - & - & - & - & - \\
    AdaTAD~\cite{Liu_2024_CVPR} & VideoMAE-H & \cmark & \xmark & \uline{88.9} & \textbf{85.3} & \uline{78.6} & \uline{66.9} & 52.5 & \uline{74.4} & - & - & - & - & - & - & - & - & - & - & - & - \\

    \rowcolor[gray]{0.9} {\textbf{\unitemploc}}  & VideoMAE-B &\cmark &\xmark & 86.1 & 81.1 & 74.6 & 63.3 & 48.8 & 70.8 & - & - & - & - & - & - & - & - & - & - & - & -\\
    \rowcolor[gray]{0.9} {\textbf{\unitemploc}}$^\dag$  & VideoMAE-B &\cmark &\xmark & 87.1 & 82.8 & 75.9 & 63.4 & 50.6 & 72.0 & - & - & - & - & - & - & - & - & - & - & - & -\\
    \rowcolor[gray]{0.9} {\textbf{\unitemploc}}  & VideoMAE-L &\cmark &\xmark & 88.8 & 84.5 & 77.9 & 66.8 & \uline{53.1} & 74.2 & - & - & - & - & - & - & - & - & - & - & - & -\\
    \rowcolor[gray]{0.9} {\textbf{\unitemploc}}$^\dag$  & VideoMAE-L &\cmark &\xmark & \textbf{89.0} & \uline{85.0} & \textbf{78.7} & \textbf{68.7} & \textbf{53.5} & \textbf{75.0} & \textbf{34.1} & \textbf{33.3} & \textbf{31.9} & \textbf{29.6} & \textbf{26.8} & \textbf{31.2} & \textbf{35.8} & \textbf{34.4} & \textbf{32.2} & \textbf{29.1} & \textbf{24.6} & \textbf{31.2} \\

    \bottomrule
\end{tabular}
\label{tab:sota_tal_simplified}
\end{table*}

\begin{table*}[t]
\centering
\caption{\textbf{Generic Event Boundary Detection Results on Kinetics-GEBD.} Results are obtained under 224$^2$ resolution.}
\vspace{-1pt}

\setlength{\tabcolsep}{8pt}
\resizebox{0.8\linewidth}{!}{

\begin{tabular}{lc|ccccccccccc}
\toprule[1.5pt]
\multirow{2}{*}{\textbf{Method}} & \multirow{2}{*}{\textbf{Backbone}} & \multicolumn{11}{c}{\textbf{F1@Rel. Dis.}}                                                                                                                                                   \\ \cmidrule{3-13} 
\multicolumn{1}{l}{}                                                 &                                & \textbf{0.05} & \textbf{0.1} & \textbf{0.15} & \textbf{0.2} & \textbf{0.25} & \textbf{0.3} & \textbf{0.35} & \textbf{0.4} & \textbf{0.45} & \multicolumn{1}{c|}{\textbf{0.5}} & \textbf{avg} \\ \midrule
BMN~\cite{bmn2019iccv}                                         & ResNet-50                                              & 18.6          & 20.4         & 21.3          & 22.0         & 22.6          & 23.0         & 23.3          & 23.7         & 23.9          & \multicolumn{1}{c|}{24.1}         & 22.3         \\
BMN-StartEnd~\cite{bmn2019iccv}                                & ResNet-50                                          & 49.1          & 58.9         & 62.7          & 64.8         & 66.0          & 66.8         & 67.4          & 67.8         & 68.1          & \multicolumn{1}{c|}{68.3}         & 64.0         \\
DDM-Net~\cite{ddmnet2022cvpr}                                     & ResNet-50                                               & 76.4          & 84.3         & 86.6          & 88.0         & 88.7          & 89.2         & 89.5          & 89.8         & 90.0          & \multicolumn{1}{c|}{90.2}         & 87.3         \\
EfficientGEBD~\cite{efficientgebd2024acmmm}                               & ResNet-50                                               & 78.3          & 85.1         & 87.4          & 88.7         & 89.6          & 90.1         & 90.5          & 90.8         & 91.1          & \multicolumn{1}{c|}{91.3}         & 86.6         \\
EfficientGEBD~\cite{efficientgebd2024acmmm}                               & ir-CSN-152-IG65M                                         & \textbf{82.9}          & 88.2         & 90.0          & 91.1         & 91.8          & 92.2         & 92.5          & 92.8         & 93.0          & \multicolumn{1}{c|}{93.2}         & 90.8         \\ \midrule
\rowcolor[gray]{0.9} \textbf{TimeLoc}                        & VideoMAE-B-Frozen                                                 &       77.5    &    86.0      &     88.8      &     90.3     &     91.3      &    92.0      &     92.4      &     92.7     &     93.0      & \multicolumn{1}{c|}{93.2}         & 89.7         \\
\rowcolor[gray]{0.9} \textbf{TimeLoc}                       & VideoMAE-B                                              & 82.0          & \underline{88.6}         & \underline{90.8}          & \underline{92.0}         & \underline{92.8}          & \underline{93.3}         & \underline{93.6}          & \underline{93.9}         & \underline{94.1}          & \multicolumn{1}{c|}{94.3}         & \underline{91.5}         \\ 
\rowcolor[gray]{0.9} \textbf{TimeLoc}                        & VideoMAE-L                                              & \underline{82.8}          & \textbf{89.1}         & \textbf{91.1}          & \textbf{92.3}         & \textbf{93.0}          & \textbf{93.5}         & \textbf{93.9}          & \textbf{94.2}         & \textbf{94.4}          & \multicolumn{1}{c|}{94.6}         & \textbf{91.9}         \\ 
\bottomrule[1.5pt]
\end{tabular}
}

\label{tab: gebd}
\vspace{-8pt}

\end{table*}

\subsection{Comparison with SoTA Methods}

\subsubsection{Temporal Action Localization}

\noindent\textbf{THUMOS14.} As shown in Table~\ref{tab:sota_tal_simplified}, TimeLoc achieves strong performance on THUMOS14, surpassing all previous end-to-end and offline methods. With the VideoMAE-B backbone, TimeLoc outperforms AdaTAD~\cite{Liu_2024_CVPR} by 0.7\% average mAP. However, AdaTAD incorporates additional adapter layers, increasing the model's capacity. 
Using the VideoMAE-L backbone (304M), we achieve an average mAP of 75.0\%, surpassing the previous best end-to-end method, AdaTAD, by 1.3\% under the same backbone and by 0.6\% even compared to the VideoMAE-H backbone (633M). This demonstrates that in temporal action localization, efficiently scaling input frames and resolution under end-to-end learning yields better performance than relying on frozen features. 
For additional results on THUMOS14, please refer to the supplementary materials.

\vspace{4pt}
\noindent\textbf{EPIC-Kitchens-100.}  
Table~\ref{tab:sota_tal_simplified} presents our results on EPIC-Kitchens-100. For both verb and noun tasks, TimeLoc, using the VideoMAE-L backbone, achieves significant performance gains over existing methods. 
With the same backbone as AdaTAD~\cite{Liu_2024_CVPR}, our approach attains an average mAP of 31.2\% on the Verb task, surpassing AdaTAD by 1.9\%. Similarly, for the Noun task, our method also achieves an average mAP of 31.2\%, outperforming previous state-of-the-art methods, including AdaTAD, by 1.9\%. These results further verify the effectiveness of the proposed unified framework under temporal action localization tasks.

\subsubsection{Generic Event Boundary Detection}
As shown in Table~\ref{tab: gebd}, our framework achieves strong performance on generic event boundary detection (GEBD) tasks using the Kinetics-GEBD dataset. 
Despite its relatively simple design, our method delivers competitive performance compared to approaches specifically designed for GEBD. It also achieves comparable F1 scores with Rel. Dis. @ 0.05. Notably, we observe a 1.1\% average F1 score improvement over the previous state-of-the-art method, \textit{i.e.}, EfficientGEBD~\cite{efficientgebd2024acmmm}, showing the effectiveness of \unitemploc.

\subsubsection{Temporal Grounding and Moment Retrieval}

In temporal grounding tasks, our method also demonstrates significant improvements over previous state-of-the-art methods. The results are presented in Table~\ref{tab:sota_charades_tacos} and ~\ref{tab:mr}.

\vspace{4pt}
\noindent\textbf{TACoS.} With the VideoMAE-B backbone, our approach achieves remarkable results with an R1@0.3 of 64.26\% and R1@0.5 of 55.79\%, outperforming the previous state-of-the-art SnAG (query-centric), which reported an R1@0.3 of 56.44\% and R1@0.5 of 44.86\%. To conduct a fair comparison, we also extracted offline features with the VideoMAE-B backbone and implemented SnAG~\cite{mu2024snag}. Using the same backbones, our proposed method demonstrates R@1 performance gains of 6.05\% and 7.78\% at tIoU=0.3/0.5 compared to the offline SnAG on the TACoS dataset. These results validate the effectiveness of our proposed end-to-end learning methodology.

\vspace{4pt}
\noindent\textbf{Charades-STA.} Our method achieves an R1@0.5 of 70.19\% and an R1@0.7 of 50.59\%, surpassing all existing methods by substantial margins of 5\% with the VideoMAE-B backbone. Additionally, there exists a performance gap of 5.51\% for R1@0.5 and 6.37\% for R1@0.7 between offline SnAG and our method using the same VideoMAE-B backbone. With the more robust VideoMAE-L backbone, our proposed method establishes new state-of-the-art results with R1@0.5 of 71.32\% and R1@0.7 of 52.96\%.

\vspace{4pt}
\noindent\textbf{QVHighlights.} As shown in Table~\ref{tab:mr}, most previous methods for moment retrieval rely on pre-extracted features and often ensemble multiple features to achieve stronger performance. However, thanks to our simple localization framework and end-to-end training, \unitemploc{} achieves an average mAP of 49.96\% with the VideoMAE-B, utilizing only 86M parameters. This surpasses the massive InternVideo2 backbone, which has 1B parameters, by a significant margin.

\vspace{4pt}
\noindent\textbf{Ego4D-NLQ.} On the recently released Ego4D dataset, our method demonstrates substantial improvements over previous state-of-the-art approaches. Using identical backbones (VideoMAE-L and CLIP text encoder), our method achieves 19.67\% R1@0.3 and 13.95\% R1@0.5, which performs 1.97\% better at R1@0.3 and 1.29\% better at R1@0.5 compared to our re-implementation of SnAG~\cite{mu2024snag}. {\unitemploc} also performs better than the recent state-of-the-art methods RGNet~\cite{hannan2024rgnet} with multiple video encoders. Detailed results on Ego4D can be found in supplementary materials.

\begin{table}[t]
\centering
\scriptsize
\caption{\textbf{Temporal Grounding Results on TACoS and Charades-STA.} $^*$ indicates that pre-extracted VideoMAE-B features are used for SnAG~\cite{mu2024snag}. The best results are in \textbf{bold}, while the second-best results are \underline{underlined}. CLIP is used as the text encoder. $^\dag$ denotes E2E methods with 224 resolution.}
\scriptsize
\setlength{\tabcolsep}{1pt}
\resizebox{1\linewidth}{!}{

\begin{tabular}{l|ccccc|ccccc}
\toprule
\multirow{3}{*}{\textbf{Method}} & 
  \multicolumn{5}{c|}{\textbf{TACoS}} &
  \multicolumn{5}{c}{\textbf{Charades-STA}} \\ \cline{2-11}
 & \multicolumn{1}{c}{\multirow{2}{*}{\textbf{Backbone}}} &
  \multicolumn{2}{c}{\textbf{R@1}} &
  \multicolumn{2}{c|}{\textbf{R@5}} &
  \multicolumn{1}{c}{\multirow{2}{*}{\textbf{Backbone}}} &
  \multicolumn{2}{c}{\textbf{R@1}} &
  \multicolumn{2}{c}{\textbf{R@5}} \\ \cline{3-6} \cline{8-11}
 & \multicolumn{1}{l}{}  & \textbf{0.3}   & \textbf{0.5}  & \textbf{0.3}   & \textbf{0.5}  & \multicolumn{1}{l}{} & \textbf{0.5}   & \textbf{0.7}   & \textbf{0.5}   & \textbf{0.7}     \\
 \midrule[0.5pt]
DRN~\cite{zeng2020drn}   & C3D & -     & 23.17 & -     & 33.36& I3D & 53.09 & 31.75 & 89.06 & 60.05  \\
CBLN~\cite{liu2021cbln}   & C3D& 38.98 & 27.65 & 73.12 & 46.24 & I3D& 61.13 & 38.22 & 90.33 & 61.69  \\
CPN~\cite{zhao2021cpn}   & C3D & 47.69 & 36.33 & -     & -     & I3D& 51.07 & 31.54 & -     & -    \\
DeNet~\cite{zhou2021denet}  & C3D  & -     & -     & -     & -    & I3D & 59.70 & 38.52 & 91.24 & 66.83   \\
MATN~\cite{zhang2021matn} & C3D   & 48.79 & 37.57 & 67.63 & 57.91 & I3D& -     & -     & -     & -    \\
VLG-Net~\cite{soldan2021vlgnet}  & C3D& 45.46 & 34.19 & 70.38 & 56.56 & I3D& -     & -     & -     & - \\
APGN~\cite{liu2021apgn}  & C3D & 40.47 & 27.86 & 59.98 & 47.12 & I3D& 62.58 & 38.86 & 91.24 & 62.11  \\
IA-Net~\cite{liu2021ianet} & C3D & 37.91 & 26.27 & 57.62 & 46.39 & I3D& 61.29 & 37.91 & 89.78 & 62.04 \\
RaNet~\cite{gao2021ranet}  & C3D  & 43.34 & 33.54 & 67.33 & 55.09 & I3D& 60.40 & 39.65 & 89.57 & 64.54 \\
MGSL-Net~\cite{liu2021ianet}  & C3D& 42.54 & 32.27 & 63.39 & 50.13 & I3D& 63.98 & 41.03 & 93.21 & 63.85 \\
MMN~\cite{wang2022mmn} & C3D& 39.24 & 26.17 & 62.03 & 47.39 & I3D& - & - & - & - \\
SSRN~\cite{zhu2023ssrn} &  C3D& 45.10 & 34.33 & 65.26 & 51.85 & I3D& 65.59 & 42.65 & \textbf{94.76} & 65.48 \\
G2L~\cite{li2023g2l} &  C3D& 42.74 & 30.95 & 65.83 & 49.86 & I3D& - & - & - & - \\
 {SnAG~\cite{mu2024snag}} & C3D        & {56.44} & {44.86} & {81.15} & {70.66} & I3D& 64.62 & {46.26} & 92.55 & {71.94} \\
 R$^2$-Tuning~\cite{liu2024r} & CLIP-B & 49.71 & 38.72 & - & - & CLIP-B & 59.78 & 37.02 & - & - \\
BAM-DETR~\cite{lee2024bam} &  SF50+CLIP-B & 56.69 & 41.54 & - & - & SF50+CLIP-B & 59.95 & 39.38 & - & - \\
{SnAG$^*$~\cite{mu2024snag}}  & VideoMAE-B        & {58.21} & {48.01} & {82.53} & {72.03} & VideoMAE-B & 64.68 & {44.22} & 90.32 & 69.19 \\

\midrule

\rowcolor[gray]{0.9} \textbf{\unitemploc}$^\dag$    & VideoMAE-B      & \uline{64.26} & \uline{55.79} & \textbf{84.98} & \textbf{77.01} & VideoMAE-B& \uline{70.19} & \uline{50.59} & 91.83 & \uline{73.04} \\
\rowcolor[gray]{0.9} \textbf{\unitemploc}$^\dag$    & VideoMAE-L      & \textbf{65.33} & \textbf{56.36} & \uline{84.38} & \uline{76.86} & VideoMAE-L & \textbf{71.32} & \textbf{52.96} & \uline{93.49} & \textbf{74.44} \\
\bottomrule
\end{tabular}
}
\label{tab:sota_charades_tacos}
\vspace{-10pt}
\end{table}

\begin{table} 
\centering
\caption{\textbf{Moment Retrieval Results on QVHighlights val subset.} $^*$ refers to E2E methods with 224 resolution.} 
\setlength{\tabcolsep}{3pt}
\scriptsize{
\begin{tabular}{l c  c  c  c}
\toprule
\textbf{Method} & \textbf{Backbone} & \textbf{mAP@0.5} & \textbf{mAP@0.75} & \textbf{Avg. mAP} \\
\midrule
Moment-DETR~\cite{lei2021detecting} & SlowFast+CLIP & - & - & 32.20 \\
UniVTG~\cite{lin2023univtg} & SlowFast+CLIP & - & - & 36.13 \\
QD-DETR~\cite{moon2023query} & SlowFast+CLIP & 62.23 & 41.82 & 41.22 \\
EaTR~\cite{jang2023knowing}   & I3D & 61.86 & 41.91 &41.74 \\
CG-DETR~\cite{moon2023correlation} & SlowFast+CLIP & 65.60 & 45.70 & 44.90 \\
CG-DETR~\cite{moon2023correlation} & InternVideo2-1B & - & -  & \underline{47.02} \\
R$^2$-Tuning~\cite{liu2024r} & CLIP & \textbf{69.04} & 47.56 & 46.17 \\
BAM-DETR~\cite{lee2024bam} & SlowFast+CLIP & 66.33 & \underline{48.22} & 46.67 \\
\midrule
\rowcolor[gray]{0.9} \textbf{\unitemploc}$^*$  & VideoMAE-B & \underline{67.58} & \textbf{52.20} & \textbf{49.96} \\
\bottomrule
\end{tabular}
}
\label{tab:mr}
\vspace{-10pt}
\end{table}

\subsection{Analysis and Findings}

In this section, we present ablation studies and additional experiments to demonstrate the key findings of our framework. Unless specified, all experiments are performed on the TACoS dataset under the temporal video grounding task.

\finding{1}{End-to-end training and scaling input frames of video encoder significantly enhance performance.} 

In Tables~\ref{tab:e2e} and~\ref{tab:scaling}, we present results with various backbones and input resolutions (both spatial resolution and temporal windows) on the TACoS dataset. All ablation studies are conducted using GloVe-6B as the text feature.

\begin{table}[t]
\scriptsize
\center
\caption{\textbf{Effectiveness of end-to-end training on TACoS.} E2E refers to end-to-end training of the video encoder. Res. refers to the spatial resolution. GloVe-6B is used to extract text feature. 
}
\setlength{\tabcolsep}{9pt}
\begin{tabular}{ccc|cc}
\toprule
\textbf{Video Enc.}   & \textbf{E2E} & \textbf{Res.}  & \textbf{R1@0.3} & \textbf{R1@0.5} \\ 
\midrule
VideoMAE-B & \xmark & 224$^2$ & 58.81      & 48.01                           \\ 
VideoMAE-B & \cmark  & 160$^2$  & \uline{62.18}  & \uline{52.14}   \\
VideoMAE-L & \xmark  & 224$^2$  &  58.56 &  48.91  \\
VideoMAE-L & \cmark & 160$^2$ & \textbf{63.26} & \textbf{53.49} \\
\bottomrule
\end{tabular}
\label{tab:e2e}
\vspace{-12pt}
\end{table}

In Table~\ref{tab:e2e}, we conduct comparative analysis between end-to-end training of the video encoder and using frozen features. With the same VideoMAE-B backbone, end-to-end joint training the video encoder can bring significant improvements of 3.37\% in R1@0.3 and 4.13\% in R1@0.5. These results highlight the enhanced capability of end-to-end training the video encoder to capture complex temporal features, thereby improving overall model accuracy.

Table~\ref{tab:scaling} examines the impact of scaling on the efficacy of the model. By increasing the spatial resolution from 160$^2$ to 224$^2$ and expanding the number of input frames, we observe considerable performance gains. Notably, the VideoMAE-L backbone exhibits marked improvement at higher resolution and extended temporal length, with R1@0.3 increasing from 57.99\% to 63.41\%, and R1@0.5 from 49.21\% to 53.86\%. These findings underscore the importance of both higher resolution and longer temporal sequences for capturing detailed spatial-temporal information essential for model performance. We also present the ablation studies on Charades-STA dataset in the supplementary materials.

\begin{table}[t]
\small 
\center
\caption{\textbf{Impact of scaling model size, spatial resolution and temporal length on TACoS.} For all these experiments in the table, we fix the text encoder.}
\setlength{\tabcolsep}{8pt}
\begin{tabular}{ccc|cc}
\toprule
\textbf{Video Enc.}   & \textbf{Res.} & \textbf{Frames} & \textbf{R1@0.3} & \textbf{R1@0.5} \\ 
\midrule
VideoMAE-B &   160$^2$  & 18,432  & 57.79  & 49.21       \\ 
VideoMAE-B &   160$^2$  & 36,864  & 62.18  & 52.14    \\ 
VideoMAE-B &   224$^2$  & 18,432  & 60.18  & 51.21    \\ 
VideoMAE-B &   224$^2$  & 36,864  & 62.68  & 53.81    \\ 
VideoMAE-L &   224$^2$  & 36,864  & \textbf{63.41}    & \textbf{53.86}     \\ 
\bottomrule
\end{tabular}
\vspace{-10pt}
\label{tab:scaling}
\end{table}

\begin{table}[t]
\caption{\textbf{Impact of different text encoders on TACoS.} Experiments are conducted under the setting of freezing both the text encoder and video encoder.}
\small
\center
\setlength{\tabcolsep}{8pt}
\begin{tabular}{ccc|cc}
\toprule
\textbf{Video Enc.} & \textbf{Text Enc.} & \textbf{E2E}          & \textbf{R1@0.3} & \textbf{R1@0.5} \\ \midrule
C3D         & GloVe-6B             & \xmark & 55.76           & 46.19           \\
C3D         & CLIP-B            & \xmark &      55.44           &   45.22    \\
C3D         & CLIP-L            & \xmark &      55.79           &   45.14    \\
C3D         & BERT-B            & \xmark &      \textbf{56.46}           &   \textbf{46.64}    \\
\bottomrule
\vspace{-15pt}
\end{tabular}

\label{tab: tacos freeze encoder}
\end{table}

\finding{2}{Unfreezing text encoder benefits feature fusion, and yields significant performance improvements on text-related tasks.} 
In Table~\ref{tab: tacos freeze encoder} and Table~\ref{tab: tacos ft ablation}, we observe that merely incorporating more advanced pre-trained text encoders results in minimal performance gains (when using the VideoMAE feature) or even performance degradation (when using the C3D feature). 
However, when we enable text encoder fine-tuning during the learning process, we observe a substantial performance improvement compared to using fixed text encoders. Table~\ref{tab: tacos ft ablation} shows that with the original VideoMAE-B feature, fine-tuning the text encoder yields 2.2\% and 2.1\% improvements on the metrics R1@0.3 and R1@0.5, respectively, comparable to the gains from fine-tuning the video encoders. When fine-tuning both text and video encoders simultaneously, our proposed method achieves better results of 64.26 and 55.79 for R1@0.3 and R1@0.5, respectively. This finding contradicts previous studies by suggesting that offline text features are suboptimal for text-related tasks.

\begin{table}[t]
\centering
\caption{\textbf{Impact of multi-stage training strategy on TACoS.} VideoMAE-B is adopted as the video encoder and 224$^2$ is used as spatial resolution of input videos.}
\small
\center
\setlength{\tabcolsep}{6.3pt}
\begin{tabular}{ccc|cc}
\toprule
 \textbf{Text Enc.}   & \textbf{E2E-\textit{Text}}       & \textbf{E2E-\textit{Video}} &        \textbf{R1@0.3}                 &     \textbf{R1@0.5}                   \\ 
\midrule
GloVe-6B                  & \xmark          & \xmark           & 58.21                   & 48.01                   \\
GloVe-6B                  & \xmark          &        \cmark     & 62.68                   & 53.81                   \\
\midrule
CLIP-B                    & \xmark          & \xmark           & 58.81                   & 49.69                   \\
CLIP-B                    &     \cmark       & \xmark           & 61.01                   & 51.79                   \\
CLIP-B                    & \xmark          &       \cmark      & 62.18                   & 52.14                   \\
CLIP-B                    &      \cmark      &         \cmark    & 64.26                   & 55.79                   \\
\midrule
BERT-B                    &       \cmark     &         \cmark    & \textbf{65.21}                   & \textbf{56.76}                   \\
\bottomrule
\end{tabular}
\label{tab: tacos ft ablation}
\vspace{-12pt}
\end{table}

\finding{3}{Our multi-stage E2E training strategy unleashes the potential of large pre-trained encoders.} 

From the results presented in Table~\ref{tab: tacos model size ablation}, we observe that in previous works, larger pre-trained models sometimes exhibit lower performance than smaller ones. This phenomenon may be attributed to the sensitivity of hyperparameters when training large models. For instance, VideoMAE-L and CLIP-L backbones perform slightly worse than VideoMAE-S/VideoMAE-B + CLIP-B feature combinations. However, within our framework, transitioning to larger pre-trained models yields consistently improved performance. The VideoMAE-B + CLIP-B configuration demonstrates a 2.20\% R1@0.3 and 1.88\% R1@0.5 performance improvement over VideoMAE-S + CLIP-B under~\unitemploc. Furthermore, when employing VideoMAE-L + CLIP-L, we observe an additional 1.07\% R1@0.3 and 0.67\% R1@0.5 performance gain compared to VideoMAE-B + CLIP-B. These results further validate the effectiveness of our proposed \unitemploc\ approach.

\vspace{4pt}
\noindent\textbf{Ablation on memories and speed.}
We present the training memory requirements and processing speed of TimeLoc on the TACoS dataset. Our analysis reveals that without the proposed temporal gradient checkpointing technique, the temporal length of the input window is limited to 576 (9,216 frames). Expanding the window size to 1152 (18,432 frames) results in out-of-memory errors. However, with temporal gradient checkpointing enabled, our approach supports window sizes of up to 2304 (36,864 frames), with potential for further expansion.
Notably, enabling temporal gradient checkpointing incurs training speed penalty. We leave speed optimization in future work.

\section{Conclusion}
\begin{table}[t]
\caption{\textbf{Impact of backbone model size on TACoS.} We use 224$^2$ as spatial resolution of input videos.}
\vspace{-10pt}
\small
\center
\setlength{\tabcolsep}{7.2pt}
\begin{tabular}{ccc|cc}
\toprule
\textbf{Video Enc.} & \textbf{Text Enc.} & \textbf{E2E} & \textbf{R1@0.3} & \textbf{R1@0.5} \\ \midrule
VideoMAE-S & CLIP-B & \xmark & 58.39 & 49.49 \\
VideoMAE-B & CLIP-B & \xmark & 58.81 & 49.69 \\
VideoMAE-L & CLIP-L & \xmark & 57.29 & 48.91 \\
\midrule
VideoMAE-S         & CLIP-B    & \cmark        & 62.06      & 53.91      \\
VideoMAE-B         & CLIP-B    & \cmark        & 64.26           & 55.79           \\
VideoMAE-L         & CLIP-L    & \cmark        & \textbf{65.33}  & \textbf{56.36}  \\ 
\bottomrule
\end{tabular}
\label{tab: tacos model size ablation}
\vspace{-3pt}
\end{table}
\begin{table}[t]
\centering
\caption{\textbf{Impact of temporal gradient checkpointing (TGC) on memory and speed on TACoS.} Speed refers to the time of one training iteration~(forward + backward). OOM means out of memory on H100-80G. Values are obtained under 224$^2$ resolution. The input frame number is equal to the window size $\times$ 16 frames.}
\vspace{-5pt             }
\small 
\center
\setlength{\tabcolsep}{3.2pt}
\begin{tabular}{ccc|cc}
\toprule
\textbf{Window Size} & \textbf{Frames} & \textbf{TGC} & \textbf{Memory$~$(GB)} & \textbf{Speed$~$(s/iter)} \\ 
\midrule
576    & 9,216   & \xmark   & 66.5  & 5.6      \\
1152    & 18,432   & \xmark   & OOM   & -       \\
\midrule
576    & 9,216   & \cmark   & 21.1  &  6.5      \\
1152    & 18,432   & \cmark   & 30.3  & 13.1        \\
2304   & 36,864   & \cmark & 46.2 & 26.3 \\
\bottomrule
\end{tabular}
\label{tab:memory cost}
\vspace{-15pt}
\end{table}

In this paper, we presented \unitemploc, a simple and accurate framework for timestamp localization in videos, including temporal action localization, temporal video grounding, generic event boundary detection, and moment retrieval. To handle long videos efficiently, we introduced temporal gradient checkpointing. Additionally, we discovered that fine-tuning a pre-trained text encoder enhances the performance of text-conditioned timestamp localization. 
The proposed \unitemploc~achieves superior accuracy and efficiency across a range of challenging benchmarks for both short and long videos. We hope our analysis on scaling and model design provides insights into video grounding and, more broadly, scalable video understanding.
{
    \small
    \bibliographystyle{ieeenat_fullname}
    \bibliography{main}
}

\clearpage
\section{Appendix}
\renewcommand\thesection{\Alph{section}}
\renewcommand\thesubsection{\thesection.\arabic{subsection}}
\setcounter{section}{0}

\section{More Implementation Details}

\begin{table}[b]
\caption{Temporal video ground results on Ego4D. $^\dag$ means result from SnAG~\cite{mu2024snag}.}
\resizebox{\linewidth}{!}{
\begin{tabular}{ccc|cc}
\toprule[1.5pt]
\textbf{Video Enc.} & \textbf{Text Enc.}    & \textbf{E2E}        & \textbf{R1@0.3}     & \textbf{R1@0.5}     \\ \bottomrule
EgoVLP$^\dag$      & EgoVLP$^\dag$     & \xmark & 15.87               & 11.26               \\
EgoVLP             & EgoVLP           & \xmark & 16.58               & 12.38               \\
CLIP + EgoVLP~\cite{hannan2024rgnet} & CLIP-B & \xmark & 18.28 & 12.04 \\
VideoMAE-L         & CLIP-B            & \xmark & 17.70                & 12.66               \\
VideoMAE-L         & CLIP-B            & \cmark & \textbf{19.67} & \textbf{13.95} \\ 
\bottomrule[1.5pt]
\end{tabular}
}
\label{tab: ego4d}
\end{table}

\begin{table}[b]
\centering
\caption{Multi-stage training strategy on TACoS. Results are obtained under 160$^2$ resolution.}
\small
\begin{tabular}{cc|cc}
\toprule[1.5pt]
\textbf{Backbone}   & \textbf{Multi Stage} & \textbf{R1@0.3} & \textbf{R1@0.5} \\ \midrule
VideoMAE-B &      \xmark       & 58.76  & 50.21  \\
VideoMAE-B &      \cmark       & 64.51  & 55.76  \\ \bottomrule[1.5pt]
\end{tabular}
\label{tab:multi_stage_ablation}
\end{table}

For the THUMOS14 dataset, we randomly sample a window of 2,304 frames with a temporal stride of 4. For EPIC-Kitchens-100, a window of 18,432 frames is randomly selected with a temporal stride of 2. On both datasets, we set the batch size to 2, and the learning rate of the head to $1\text{e}^{-4}$. 

For the Kinetics-GEBD dataset, the temporal stride is set to 1, and we set the batch size to 16 and use learning rates of $1\text{e}^{-5}$ and $1\text{e}^{-3}$ for the video encoder and head, respectively.

For temporal grounding datasets, we extract a window of 2,304 frames with a temporal stride of 4 for Charades-STA and a window of 36,864 frames with a temporal stride of 4 for TACoS and Ego4D. For the QVHighlights dataset, we set the frame rate to 8 FPS and use all available frames (approximately 1,200 frames). The frame resolution is standardized at $160^2$ across all experiments.
In Stage 1, we set the learning rate ratio of the text encoder to other modules to 1:10 and use a base learning rate of $1\text{e}^{-3}$. For Ego4D, we maintain the same learning rate ratio but reduce the base learning rate to $1\text{e}^{-4}$. In Stage 2, we freeze the text encoder and transfer its learning rate to the video encoder. We also assign a lower learning rate for the text adapter in the fusion block, specifically $2\text{e}^{-5}$ for Ego4D and $2\text{e}^{-4}$ for other datasets.
In all stages, we use a default batch size of 4. However, for Ego4D, the batch size is set to 2, and for QVHighlights, it is set to 8. 

For post-processing, we apply top-$k$ selection followed by SoftNMS and segment voting to refine the predicted segments. We set the top-$k$ value to 100 for temporal grounding and generic event boundary detection, and to 2000 for temporal action localization and highlight detection.

\section{Ablations on Charades-STA}
We further present ablation results with different scaling parameters but consistent text encoder configurations on the Charades-STA dataset in Table~\ref{tab: charades scaling}. Despite the distinct characteristics between TACoS (long-duration videos with numerous queries) and Charades-STA (shorter 30-second videos with fewer queries), our proposed method demonstrates similar performance improvements across both datasets. End-to-end training yields approximately 6\% performance improvement on R1@0.5 and R1@0.7. Additionally, scaling the spatial resolution and backbone architecture provides a further 1-2\% performance enhancement. When utilizing VideoMAE-L as the backbone, we achieve a new state-of-the-art performance with R@1 of 71.32 and 52.96 at tIoU thresholds of 0.5 and 0.7, respectively.

\begin{table}[b]
\caption{Ablation studies on Charades-STA. E2E means end-to-end training, and Frozen refers to freezing the backbone during training.}
\resizebox{\linewidth}{!}{
\begin{tabular}{ccccc|cc}
\toprule[1.5pt]
\textbf{Method} & \textbf{Backbone}          & \textbf{E2E}      & \textbf{Frozen}                 & \textbf{Res.}                  & \textbf{R1@0.5} & \textbf{R1@0.7} \\ \bottomrule
SnAG   & C3D               & \xmark  & \cmark   & 224$^2$ & 51.75  & 47.96  \\ 
SnAG   & I3D               & \xmark  & \cmark   & 224$^2$ & 65.19  & 46.32  \\ 
\unitemploc   & VideoMAE-B        & \xmark  & \cmark   & 224$^2$ & 64.6   & 44.2   \\ 
\unitemploc   & VideoMAE-B & \cmark & \cmark & 224$^2$ & 67.28  & 47.2   \\
\unitemploc   & VideoMAE-B        & \cmark & \xmark & 224$^2$ & 70.19  & 50.48  \\ 
\unitemploc   & VideoMAE-B        & \cmark & \xmark & 336$^2$ & \uline{70.99}  & \uline{52.85}  \\ 
\unitemploc   & VideoMAE-L        & \cmark & \xmark & 224$^2$ & \textbf{71.32}  & \textbf{52.96}  \\ 
\bottomrule[1.5pt]
\end{tabular}
}

\label{tab: charades scaling}
\end{table}

\begin{table*}[t]
\centering
\caption{\textbf{General Event Boundary Detection Results on Kinetics-GEBD.} Results are obtained under 224$^2$ resolution.}
\vspace{-1pt}

\setlength{\tabcolsep}{4pt}
\resizebox{0.7\linewidth}{!}{

\begin{tabular}{lcc|ccccccccccc}
\toprule[1.5pt]
\multirow{2}{*}{\textbf{Method}} & \multirow{2}{*}{\textbf{Backbone}} & \multirow{2}{*}{\textbf{Res}} & \multicolumn{11}{c}{\textbf{F1@Rel. Dis.}}                                                                                                                                                   \\ \cmidrule{4-14} 
\multicolumn{1}{l}{}                                                 &                                & & \textbf{0.05} & \textbf{0.1} & \textbf{0.15} & \textbf{0.2} & \textbf{0.25} & \textbf{0.3} & \textbf{0.35} & \textbf{0.4} & \textbf{0.45} & \multicolumn{1}{c|}{\textbf{0.5}} & \textbf{avg} \\ 
\midrule
\textbf{TimeLoc}                        & VideoMAE-B-Frozen           & 224$^2$                                      &       77.5    &    86.0      &     88.8      &     90.3     &     91.3      &    92.0      &     92.4      &     92.7     &     93.0      & \multicolumn{1}{c|}{93.2}         & 89.7         \\
\textbf{TimeLoc}                        & VideoMAE-B                 & 160$^2$                              & 81.7          & 88.4         & 90.7          & 91.9         & 92.7          & 93.2         & 93.5          & 93.8         & 94.0          & \multicolumn{1}{c|}{94.2}         & 91.4         \\
\textbf{TimeLoc}                       & VideoMAE-B                 & 224$^2$                             & 82.0          & 88.6         & 90.8          & 92.0         & 92.8          & 93.3         & 93.6          & 93.9         & 94.1          & \multicolumn{1}{c|}{94.3}         & 91.5 \\
\bottomrule[1.5pt]
\end{tabular}
}
\label{tab: gebd_res_ablation}
\vspace{-3pt}
\end{table*}

\section{Ablations on Ego4D}

We also present results on Ego4D-NLQ in Table~\ref{tab: ego4d}. The original results of SnAG~\cite{mu2024snag} achieve only 15.87\% R1@0.3 and 11.26\% R1@0.5. For a fair comparison, we first re-implemented SnAG within our framework, obtaining 16.58\% R1@0.3 and 12.38\% R1@0.5, which is slightly higher than the originally reported results. 

Next, we replaced the video backbone with VideoMAE-L and the text encoder with CLIP-B, leading to an additional performance boost of approximately 1\% and 0.3\%, respectively. TimeLoc further improves upon the offline baseline, achieving gains of 1.97\% R1@0.3 and 1.29\% R1@0.5. Moreover, TimeLoc outperforms the recent state-of-the-art RGNet~\cite{hannan2024rgnet}, which utilizes two video encoders, further demonstrating the effectiveness of our approach.

\section{Ablations on Kinetics-GEBD}
In the General Event Boundary Detection task, we also explored the impact of spatial resolution. As shown in Table~\ref{tab: gebd_res_ablation}, increasing the resolution from 160$^2$ to 224$^2$ only leads to a minor improvement of 0.3 in F1@0.05 and 0.1 in F1@avg, which is significantly smaller than the impact brought by end-to-end training.

\section{Ablations on Multi-Stage Training}

In Section 3.4 of the main paper, we propose a multi-stage training strategy for fine-tuning the text encoder. Compared to directly performing end-to-end fine-tuning of both the video and text encoders simultaneously, our approach significantly reduces learning difficulty and hyperparameter sensitivity, leading to improved performance. 

As shown in Table~\ref{tab:multi_stage_ablation}, jointly fine-tuning the video and text encoders achieves only 58.76 R1@0.3 and 50.21 R1@0.5, exhibiting a substantial gap compared to the multi-stage training approach.

\begin{table}[t]
\centering
\scriptsize
\caption{Temporal Action Localization Results on the THUMOS14 Dataset. We report mAP (\%) at different tIoUs. E2E denotes end-to-end training, while Flow refers to offline extracted optical flow features. The best results are in \textbf{bold}, while previous best results are \underline{underlined}. $^\dag$ indicates results obtained under a 224$^2$ spatial resolution.}
\scriptsize
\setlength{\tabcolsep}{1.5pt}
\resizebox{1\linewidth}{!}{
\begin{tabular}{l|ccc|ccccc>{\columncolor[gray]{0.9}}c}
\toprule
\multirow{2}{*}{\textbf{Method}} & \multicolumn{1}{c}{\multirow{2}{*}{\textbf{Backbone}}} & \multicolumn{1}{c}{\multirow{2}{*}{\textbf{E2E}}} & \multirow{2}{*}{\textbf{Flow}} & \multicolumn{6}{c}{\textbf{THUMOS-14}} \\ \cline{5-10} 
                   &               &            &                       & \textbf{0.3} & \textbf{0.4} & \textbf{0.5} & \textbf{0.6} & \textbf{0.7} & \textbf{Avg.}\\
\hline

BMN~\cite{lin2019bmn}&TSN&\xmark&\cmark&{56.0} & 47.4 & 38.8 & 29.7 & 20.5 & 38.5 \\

TadTR~\cite{liu2022end}  & I3D &\xmark&\cmark& 62.4 & 57.4 & 49.2 & 37.8 & 26.3 & 46.6 \\

ActionFormer~\cite{zhang2022actionformer} &SlowFast-R50&\xmark&\xmark& 78.7 & 73.3 & 65.2 & 54.6 & 39.7 & 62.3 \\
ActionFormer~\cite{zhang2022actionformer} &I3D&\xmark&\cmark&82.1 & 77.8 & 71.0 & 59.4 & 43.9 & 66.8\\

ASL~\cite{shao2023action}  & I3D &\xmark &\cmark& 83.1 & 79.0 & 71.7 & 59.7 & 45.8 & 67.9  \\
TriDet~\cite{shi2023tridet} & I3D &\xmark &\cmark& 83.6 & 80.1 & 72.9 & 62.4 & 47.4 & 69.3 \\
VideoMAEv2~\cite{wang2023videomae}  & VideoMAEv2-g &\xmark &\xmark& - & - & - & - & - & 69.6 \\
InternVideo~\cite{wang2022internvideo}  & \scriptsize{VideoMAE-H*} &\xmark &\xmark& - & - & - & - & - & \underline{71.5} \\
DyFaNet~\cite{yang2024dyfadet}  & \scriptsize{VideoMAE-g} & \xmark & \xmark &84.3 & - & 73.7 & - & 50.2 & 70.5 \\ 
DyFaNet~\cite{yang2024dyfadet}  & \scriptsize{VideoMAE-g} & \xmark & \cmark &85.4 & - & 74.0 & - & 50.2 & 71.1 \\ 

\hline
\hline

AFSD~\cite{lin2021learning} &I3D&\cmark &\cmark&67.3 &62.4 &55.5 & 43.7 & 31.1 & 52.0 \\
E2E-TAD~\cite{liu2022empirical} & SlowFast-R50 &\cmark &\xmark & 69.4 & 64.3 & 56.0 & 46.4 & 34.9 & 54.2 \\
BasicTAD~\cite{yang2023basictad} & SlowOnly-R50 &\cmark &\xmark& 75.5 & 70.8 & 63.5 & 50.9 & 37.4 & 59.6\\
TALLFormer~\cite{cheng2022tallformer}&VideoSwin-B &\cmark &\xmark&76.0 &-&63.2&-& 34.5 & 59.2 \\
Re$^2$TAL~\cite{zhao2023re2tal} &Re$^2$VideoSwin-T &\cmark &\xmark& 77.0 & 71.5 & 62.4 & 49.7 & 36.3 & 59.4 \\
AdaTAD~\cite{Liu_2024_CVPR} & VideoMAE-B & \cmark & \xmark & 87.0 & 82.4 & 75.3 & 63.8 & 49.2 & 71.5 \\
AdaTAD~\cite{Liu_2024_CVPR}$^\dag$ & VideoMAE-B & \cmark & \xmark & - & - & - & -& - & 71.9 \\
AdaTAD~\cite{Liu_2024_CVPR} & VideoMAE-L & \cmark & \xmark & 87.7 & 84.1 & 76.7 & 66.4 & 52.4 & 73.5 \\
AdaTAD~\cite{Liu_2024_CVPR}$^\dag$  & VideoMAE-L & \cmark & \xmark & - & - & -& - & - & 73.7 \\
AdaTAD~\cite{Liu_2024_CVPR} & VideoMAE-H & \cmark & \xmark & \uline{88.9} & \textbf{85.3} & \uline{78.6} & \uline{66.9} & 52.5 & \uline{74.4} \\
\rowcolor[gray]{0.9} {\textbf{\unitemploc}}  & VideoMAE-B &\cmark &\xmark & 86.1 & 81.1 & 74.6 & 63.3 & 48.8 & 70.8 \\
\rowcolor[gray]{0.9} {\textbf{\unitemploc}}$^\dag$  & VideoMAE-B &\cmark &\xmark & 87.1 & 82.8 & 75.9 & 63.4 & 50.6 & 72.0 \\
\rowcolor[gray]{0.9} {\textbf{\unitemploc}}  & VideoMAE-L &\cmark &\xmark & 88.8 & 84.5 & 77.9 & 66.8 & \uline{53.1} & 74.2\\
\rowcolor[gray]{0.9} {\textbf{\unitemploc}}$^\dag$  & VideoMAE-L &\cmark &\xmark & \textbf{89.0} & \uline{85.0} & \textbf{78.7} & \textbf{68.7} & \textbf{53.5} & \textbf{75.0} \\
\bottomrule
\end{tabular}
}
\label{tab:sota_thumos}
\end{table}

\section{Additional Results on THUMOS14}
In Table~\ref{tab:sota_thumos}, we show more results on the THUMOS dataset for comparison. The results demonstrate that {\unitemploc}~achieves state-of-the-art performance compared to both offline and E2E methods, even when some previous methods utilize stronger backbones (e.g., VideoMAE-g) or incorporate additional optical flow inputs.

\clearpage

\end{document}